# Merging uncertain knowledge bases in a possibilistic logic framework

**Salem Benferhat**
IRIT-UPS
118, Route de Narbonne
31062 Toulouse Cedex, France
e-mail: benferhat@irit.fr

**Claudio Sossai**
Ladseb-CNR
Corso Stati Uniti 4
35127 Padova PD, Italy
sossai@ladseb.pd.cnr.it

## Abstract

This paper addresses the problem of merging uncertain information in the framework of possibilistic logic. It presents several syntactic combination rules to merge possibilistic knowledge bases, provided by different sources, into a new possibilistic knowledge base. These combination rules are first described at the meta-level outside the language of possibilistic logic. Next, an extension of possibilistic logic, where the combination rules are inside the language, is proposed. A proof system in a sequent form, which is sound and complete with respect to the possibilistic logic semantics, is given.

## 1 Introduction

Merging uncertain information is a crucial problem in designing knowledge base systems. In many situations, relevant information is provided by different sources. This requires to perform some combination operation which uses simultaneously the information provided by the different sources in order to come to a conclusion. The way this problem is tackled depends on the way the information is represented. On the one hand, pieces of information pertaining to numerical parameters are usually represented by distribution functions (in the sense of some uncertainty theory). These distributions are directly combined by means of operations and yield a new distribution. On the other hand, information may be also expressed in logical terms, which may be, however, pervaded with uncertainty. In this case, some uncertainty weights are attached to the logical formulae. Although similar issues are raised in the two frameworks, like the handling of conflicting information, the two lines of research in numerical data fusion (e.g., [Abidi and Gonzalez 1992, Flamm and Luisi 1992]) and in symbolic combination (e.g., [Cholvy1992, Benferhat et al. 1995]) have been investigated independently.

This paper studies the parallel combination of uncertain knowledge bases. Uncertainty is represented in the possibility theory framework either at the semantical level via possibility distributions or at the syntactical level using possibilistic formulae. Each set of possibilistic formulae can be represented by a possibility distribution and conversely. The combination of possibility distributions has been studied in [Dubois and Prade 1992], and different combination methods have been proposed. In this paper, we apply these combination methods to the possibility distributions associated with the sets of weighted formulae and we look for the syntactical counterparts of these combinations on the sets of possibilistic formulae.

The rest of the paper is organized as follows: Section 2 recall the basic elements of standard possibilistic logic (SPL) needed to read the paper (for a complete exposition on SPL; see [Dubois et al. 1994]). Section 3 recalls the semantical combination rules developed in [Dubois and Prade 1992], and proposes their syntactical counterpart applied to the possibilistic knowledge bases. Section 4 introduces an extension of possibilistic logic, that we call LPL, where the language is enriched with two new connectives representing a new negation and a new conjunction. An example, inspired from an application in mobile robotics, is, also given.

## 2 Standard Possibilistic Logic SPL

### 2.1 Possibility distribution and possibilistic entailment

In this section, we only consider a propositional language. Greek letters $\alpha, \beta,...$ represent real numbers belonging to [0,1], and uppercase latin letters (A, B,...) represent propositional formulae. Let $\mathcal{W}$ be the set of interpretations, one of them being the actual real world. A possibility distribution is a mapping $\pi$ from $\mathcal{W}$ to the interval [0,1]. $\pi$ is said to be normalized if $\exists \omega \in \mathcal{W}$, such that $\pi(\omega) = 1$. $\pi$ represents some background knowledge; $\pi(\omega) = 0$ means that the state $w$ is impossible, and $\pi(\omega) = 1$ means that nothing prevents $w$ from being the real world. When $\pi(w_1) > \pi(w_2)$, $w_1$ is a preferred candidate to $w_2$ as the real state of the world. A possibility distribution $\pi$ induces a mapping grading the necessity (or certainty) of a formula $A$ which evaluates to what extent $A$ is entailed by the available knowledge. The necessity measure Nec is defined by: $Nec_\pi(A) = 1 - \max\{\pi(\omega) \mid \omega \models \neg A\}$.

**Def. 1** *A weighted formula (B α) is said to be a plausible conclusion of $\pi$, denoted by $\pi \Vdash (B\ \alpha)$, if and only if i) $Nec_\pi(B) > Nec_\pi(\neg B)$, and ii) $Nec_\pi(B) \geq \alpha$.*

If $\pi$ is normalized, then (ii) implies (i). When (B α) is a conclusion of $\pi$, we say that $\pi$ forces (B α).

### 2.2 From possibilistic knowledge bases to possibility distributions

A possibilistic knowledge base is made of a finite set of weighted formulae $\Sigma = \{(A_i\ \ \alpha_i), i = 1, n\}$ where $\alpha_i$ is understood as a lower bound on the degree of necessity $Nec(A_i)$. A possibilistic knowledge base $\Sigma$ can be associated with a semantics in terms of possibility distributions. $\pi$ is said to be compatible with $\Sigma$ if for each $(A_i\ \ \alpha_i) \in \Sigma$, we have : $Nec_\pi(A_i) \geq \alpha_i$. In general, there are several possibility distributions compatible with $\Sigma$. One way to select one possibility distribution is to use the minimum specificity principle. A possibility distribution $\pi$ is said to be less specific (or less informative) than $\pi_1$ if $\forall \omega, \pi(\omega) \geq \pi_1(\omega)$. The minimum specificity principle allocates to the interpretations the greatest possibility degrees in agreement with the constraints $Nec_\pi(A_i) \geq \alpha_i$. We denote by $\pi_\Sigma$ the least specific possibility distribution which is compatible with $\Sigma$. $\pi_\Sigma$ is defined by (e.g., [Dubois et al. 1994])
$$\forall w \in \mathcal{W}, \pi_\Sigma(w) = min_{i=1,n}\{1 - \alpha_i, w \models \neg A_i\}.$$
If $w$ satisfies all the formulae in $\Sigma$ then $\pi_\Sigma(w) = 1$. The possibility distribution $\pi_\Sigma$ is not necessarily normal, and $Inc(\Sigma) = 1 - max_{w \in \mathcal{W}} \pi_\Sigma(w)$ is called the degree of inconsistency of the knowledge base $\Sigma$.

### 2.3 Possibilistic inference

The possibilistic logic inference can be performed at the syntactical level by means of a weighted version of the resolution principle:
$$\frac{(A \vee B\ \ \ \alpha)\quad (\neg A \vee C\ \ \ \beta)}{(B \vee C\ \ \ min(\alpha, \beta))}$$
It has been shown that $Inc(\Sigma)$ corresponds to the greatest lower bound that can be obtained for the empty clause by the repeated use of the above resolution rule. Proving (A α) from a possibilistic knowledge base $\Sigma$ comes down to deriving the contradiction $(\perp\ \beta)$ from $\Sigma \cup \{(\neg A\ \ 1)\}$ with a weight $\beta \geq \alpha > Inc(\Sigma)$. It will be denoted by $\Sigma \vdash_{pref} (A\ \ \alpha)$. This inference method is sound and complete with respect to the possibilistic semantics. Namely [Dubois et al. 1994]:
$\pi_\Sigma \Vdash (A\ \ \alpha)$ iff $\Sigma \vdash_{pref} (A\ \ \alpha)$ with $\alpha > Inc(\Sigma)$.
The syntactic inference $\vdash_{pref}$ is efficient and has a complexity similar to that of classical logic.

## 3 Merging uncertain information

In [Dubois and Prade 1992], several proposals have been made to address the problem of combining $n$ possibility distributions $\pi_{1,n}$ into a new possibility distribution. All the proposed combination modes are defined at the semantical level. This means that these semantical combinations are impractical for large languages. Therefore, we need to look for syntactical combination rules which operate directly on possibilistic knowledge bases. Let us first recall the semantical combination methods [Dubois and Prade 1992], and then we investigate their syntactical counterparts.

### 3.1 Merging possibility distributions

The basic combination modes in the possibilistic setting are the minimum and the maximum of possibility distributions. Namely define:
$$\forall \omega, \quad \pi_{cm}(\omega) = min_{i=1,n} \pi_i(\omega),$$
$$\forall \omega, \quad \pi_{dm}(\omega) = max_{i=1,n} \pi_i(\omega).$$
The conjunctive aggregation makes sense if all the sources are regarded as equally and fully reliable since all values that are considered as impossible by one source but possible by the others are rejected, while the disjunctive aggregation corresponds to a weaker reliability hypothesis where there is at least one reliable source for sure, but we do not know which one.
The previous combination modes based on maximum and minimum operators have no reinforcement effect. Namely, if expert 1 assigns possibility $\pi_1(\omega) < 1$ to interpretation $\omega$, and expert 2 assigns possibility $\pi_2(\omega) < 1$ to this interpretation then overall, in the conjunctive mode, $\pi(\omega) = \pi_1(\omega)$ if $\pi_1(\omega) < \pi_2(\omega)$, regardless of the value of $\pi_2(\omega)$. However since both experts consider $\omega$ as rather impossible, and if these opinions are independent, it may sound reasonable to consider $\omega$ as less possible than what each of the experts claims. This type of combination cannot be modelled by any idempotent operation, but can be obtained using a triangular norm operation other than min, and a triangular conorm operation other than max.

**Def. 2** *A triangular norm (for short t-norm) $\otimes$ is a two place function whose domain is the unit square [0,1]X[0,1] and which satisfies the conditions:*
*1. $0 \otimes 0 = 0$, $\alpha \otimes 1 = 1 \otimes \alpha = \alpha$;*
*2. $\alpha \otimes \beta \leq \delta \otimes \gamma$ whenever $\alpha \leq \delta$ and $\beta \leq \gamma$ ;*
*3. $\alpha \otimes \beta = \beta \otimes \alpha$;*
*4. $\alpha \otimes (\beta \otimes \delta) = (\alpha \otimes \beta) \otimes \delta$.*

A triangular conorm (for short t-conorm) $\oslash$ is a two place function in the unit square [0,1] x [0,1] which satisfies the conditions 2-4 given in the previous definition plus the following conditions:
5. $1 \oslash 1 = 1$, $\alpha \oslash 0 = 0 \oslash \alpha = \alpha$.
Any t-conorm $\oslash$ can be generated from a t-norm $\otimes$ through the duality transformation:
$$\alpha \oslash \beta = 1 - (1-\alpha \otimes 1-\beta).$$
and conversely. The basic t-norms are the minimum operator, the product operator and the t-norm $max(0, \alpha + \beta - 1)$ called "Lukasiewicz t-norm". The duality relation respectively yields the following t-conorms: the maximum operator, the "probabilistic sum" $\alpha + \beta - \alpha * \beta$, and the "bounded sum" $min(1, \alpha + \beta)$. We shall denote by $\pi_\otimes$ and $\pi_\oslash$ the possibility distributions resulting from the combination using a t-norm $\otimes$ and a t-conorm $\oslash$ respectively.

### 3.2 Syntactical combination modes

We are interested in the combination of n possibilistic knowledge bases $\Sigma_{i=1,n}$ provided by n sources. Each knowledge base $\Sigma_i$ is associated with a possibility distribution $\pi_i$ which is its semantical counterpart. Given a semantical combination rule $C_{sem}$ on the $\pi_i$'s, reviewed in the previous section, we look for a syntactical combination $C_{syn}$ on the $\Sigma_i$'s such that:
$$\pi_{C_{sem}} \Vdash (B, \alpha) \text{ iff } \Sigma_{C_{syn}} \vdash_{pref} (B, \alpha)$$
with $\alpha > Inc(\Sigma_{C_{syn}})$, and where $\pi_{C_{sem}}$ (resp. $\Sigma_{C_{syn}}$) is the possibility distribution (resp. the knowledge base) obtained by merging the $\pi_i$'s (resp. the $\Sigma_i$'s) using $C_{sem}$ (resp. $C_{syn}$). Let us first consider the simple case of combining two one-formula knowledge bases $\Sigma_1 = \{(A \quad \alpha)\}$ and $\Sigma_2 = \{(B \quad \beta)\}$. Let $\pi_1$ and $\pi_2$ be the possibility distributions associated respectively to $\Sigma_1$ and $\Sigma_2$. We are looking for a knowledge base $\Sigma_\otimes$ which associated possibility distribution is equal to $\pi_\otimes$. Then we can check that $\Sigma_\otimes$ results in three possibilistic formulae: $\Sigma_\otimes = \Sigma_1 \cup \Sigma_2 \cup \{(A \vee B \quad \alpha \oslash \beta)\}$, where $\oslash$ is the t-conorm dual to $\otimes$. The generalisation to the case of any knowledge bases and also to any t-norm or t-conorm is given by the following theorem:

**Theorem 1** Let $\Sigma_1 = \{(A_i \quad \alpha_i) | i \in I\}$ and $\Sigma_2 = \{(B_j \quad \beta_j) | j \in J\}$ be two bases. Let $\pi_1$ and $\pi_2$ be their associated possibility distributions. Then, $\pi_\otimes$ and $\pi_\oslash$ are associated with the following bases:
$\Sigma_\otimes = \Sigma_1 \cup \Sigma_2 \cup \{(A_i \vee B_j \quad (\alpha_i \oslash \beta_j)) | i \in I \text{ and } j \in J\}$
(where $\oslash$ is the t-conorm dual to the t-norm $\otimes$)
$\Sigma_\oslash = \{(A_i \vee B_j \quad \alpha_i \otimes \beta_j) | i \in I \text{ and } j \in J\}$
(where $\otimes$ is the t-norm dual to the t-conorm $\oslash$).

The proof can be found in (Benferhat et al., 1997). $\Sigma_\otimes$ may be inconsistent, but however $\Sigma_\oslash$ is always consistent (provided that $\Sigma_1$ or $\Sigma_2$ is consistent). In order to reduce the size of $\Sigma_\otimes$ and $\Sigma_\oslash$, the following definitions and facts are useful:

**Def. 3** : A possibilistic formula $(A \; \alpha)$ of $\Sigma$ is said to be subsumed by $\Sigma$ iff $Cut(\Sigma, \alpha) \vdash A$ where $Cut(\Sigma, \alpha) = \{B; (B \quad \beta) \in \Sigma - \{(A \quad \alpha)\} \text{ and } \beta \geq \alpha\}$.

**Fact 1** Let $(A \; \alpha) \in \Sigma$ be a subsumed formula. Then $\Sigma$ and $\Sigma - \{(A \quad \alpha)\}$ are equivalent, i.e., they induce the same possibility distribution.

**Fact 2** The two knowledge bases $\Sigma$ and $\Sigma - \{(T \quad \alpha)\}$, where $T$ denotes tautology, are equivalent.

On the basis of the previous facts we get:

**Fact 3** Let $\Sigma_1$ and $\Sigma_2$ be two knowledge bases. Let $\Sigma'_1$ be a knowledge base obtained from $\Sigma_1$ by removing subsumed beliefs and tautologies. Then combining $\Sigma_2$ with $\Sigma_1$ is equivalent to combining $\Sigma_2$ with $\Sigma'_1$.

### 3.3 Example

Before giving a short example, we briefly clarify which operators should be used and in which order the information should be merged. When combining several bases with a same t-norm (resp. a same t-conorm) then the order has no importance, however when several t-norms or t-cornorms are simultaneously used then associativity is no longer guaranteed. Indeed, for instance $min(\pi_1, \pi_2) * \pi_3 \neq \pi_1 * min(\pi_2, \pi_3)$. The choice of the operator is related to the dependencies of the sources. If the sources are independent then it is recommended to use a t-norm different from the minimum in order to get some reinforcement effect. Moreover, if the pieces of information are not conflicting then it is preferred to use a t-norm, otherwise the maximum is more suitable. In practice, a t-conorm different from the maximum is rarely recommended since it yields a cautious inference, and hence a loss of information. Recall that: $\alpha \oslash \beta \geq max(\alpha, \beta)$.

In general, the available data are:
1. A set of hard rules $\Sigma_h$ which expresses some integrity constraints. These rules have certainty degrees equal to 1 and are available to all the sources,
2. A set of uncertain knowledge bases $\Sigma_i$ which comes from different sources, and
3. A set of facts $F_j$ which focuses on particular situations. This set of facts can be common to all the sources or specific to each source (in this case, each source provide a knowledge base and a set of facts).

We can imagine different schemas to merge these information:
1. Combine all the information, with a same t-norm (resp. t-conorm), or
2. Apply (i.e., add) the facts to each base, then combine the results with a t-norm (resp. t-conorm), or
3. Combine first the possibilistic knowledge bases then add the facts to the resulting knowledge base, or
4. Proceed locally (and recursively) by first applying the facts to some bases, then combine the results by some t-norm (resp. t-conorm). Next, use the result of this combination as an input to other bases, which will again be merged by other t-norms (resp. t-conorms). Repeat iteratively this operation until the request is reached or all the bases and facts are used.

Clearly, the use of a given schema of combination depends on the considered application. In the example considered in this paper, the last schema is preferred. The simplified example considered in this section concerns the problem of position estimation (localization) of a mobile robot in a partially and approximately known indoor environment. The environment is described as a graph of "relevant places" $p_1, \ldots, p_n$ connected by paths. Each place $p_i$ corresponds to a corner, a door, a section of a corridor etc., and is characterized by a "sensory signature" that captures its distinctive features. The signature is the result of two sources of sensory information:
. a ring of ultrasonic sensors which allow to detect the presence of walls or, more generally, of occlusions
. a video camera which tracks a fixed light beacon and makes it possible to associate each place with a different image of the beacon itself. For a complete description, see [Bison et al., 1997a,b,1998].

Suppose we are interested in knowing if the robot is at the North-West corner of a given room. We have two different knowledge bases expressing the information provided by the two independent sources (sonar sensors and camera). The following notations are used:
$A = sonar(wall, west), B = sonar(occlusion, north),$
$C = sonar(occlusion, west), D = sonar(corner, north, west),$
$E = close(wall, west), F = close(wall, north),$
$G = cameralocation - at(p_1), H = at(corner, north, west).$
The two bases associated to the sources are:
. $\Sigma_s = \{(D \wedge E \wedge F \rightarrow H \quad 1), (B \wedge C \rightarrow D \quad .5), (A \wedge B \rightarrow D \quad .8)\}$ where $\rightarrow$ is the material implication.
The rules mean:
- If a corner is perceived in the North-West direction, and if the robot is close to the walls at North and West, then we can say that the robot is at the corner;
- If occlusions are perceived in the two directions North-West, then we can say that a corner is perceived at North-West with certainty degree .5;
- If an occlusion is perceived at North, and a wall is perceived at West, then we can say that a corner is perceived with a higher certainty degree .8.
. $\Sigma_c = \{(G \rightarrow H \quad 1)\}$
This rule means that if the camera sees a corner in the North-West direction, then we can say that the robot is at the North-West corner.

We also assume that we have the two sets of facts:
$Fact_s = \{(A \quad .4), (B \quad .5), (C \quad .8), (E \quad .7), (F \quad .4)\}$
$Fact_c = \{(G \quad .6)\}$
Now, let us see to what degree we can deduce that the robot is at the corner, and let us see the influence of the choice of the combination schema. Note that all the available information is consistent, therefore we use a t-norm rather than a t-conorm for merging.
1. The first case is to use the minimum operator, namely we take the concatenation of all the bases and facts. We denote by $\Sigma_f$ the final base. To check if the robot is at the corner, we first add $(\neg H \; 1)$ to $\Sigma_f$. Then applying the possibilistic resolution between $(G \rightarrow H \quad 1)$ and $(\neg H \quad 1)$ leads to have $(\neg G \quad 1)$, and from $(\neg G \quad 1)$ and $(G \quad .6)$ we get a contradiction to a degree .6. We can check that .6 is the best lower bound, since the set of information in $\Sigma_f$ having certainty degree strictly higher than .6 is consistent with $(\neg H \quad 1)$. Hence, using the minimum leads to infer that the robot is at corner to a degree .6.
2. The second case consists in first combining the bases $\Sigma_s$ and $\Sigma_c$ with some t-norm $\otimes$, then adding the facts to the resulting base. We can easily check that this leads to the same conclusion as in the first case. The reason is that combining $\Sigma_s$ and $\Sigma_c$ with $\otimes$ (even different from the minimum) leads to a knowledge base equivalent to $\Sigma_s \cup \Sigma_c$. This is due to the fact that the certainty degree of the rule in $\Sigma_c$ is equal to 1 (hence, all the added beliefs will be subsumed by this rule).
3. The last case concerns a local merging. We first add the set $Fact_s$ to $\Sigma_s$. We denote $\Sigma_{fs} = Fact_s \cup \Sigma_s$ the result of this step. We do the same thing with the second base, and we denote $\Sigma_{fc} = Fact_c \cup \Sigma_c$. Since the two sources are independent, we use a t-norm different from the minimum. Let $\Sigma_{f*}$ be the base resulting of combining $\Sigma_{fc}$ and $\Sigma_{fs}$ using the product t-norm. We have: $\Sigma_{f*} = \Sigma_{fs} \cup \Sigma_{fc} \cup \{(B \wedge C \rightarrow$

$D \wedge G \quad .8), (A \wedge B \rightarrow D \wedge G \quad .92), (A \vee G \quad .76), (B \vee G \quad .8), (C \vee G \quad .92), (E \vee G \quad .88), (F \vee G \quad .76)\}$
Now, let us add to $\Sigma_{f*}$ the assumption $(\neg H \quad 1)$. Then applying the resolution to $(G \rightarrow H \quad 1)$ and $(\neg H \quad 1)$ leads to have $(\neg G \quad 1)$. Now, we can check that applying all the possible resolutions between $(\neg G \quad 1)$ and the rules: $\{(B \wedge C \rightarrow D \wedge G \quad .8), (A \wedge B \rightarrow D \wedge G \quad .92), (A \vee G \quad .76), (B \vee G \quad .8), (C \vee G \quad .92), (E \vee G \quad .88), (F \vee G \quad .76)\}$ leads to get the formulae: $\{(D \quad .8), (E \quad .88), (F \quad .76)\}$. Applying again the resolution of these formulae with the rule: $(D \wedge E \wedge F \rightarrow H \quad 1)$ leads to infer $(H \quad .76)$, which contradicts the assumption $(\neg H \quad 1)$ to a degree .76. The degree .76 is the best lower bound. Note that, if we use the "Lukasiewicz" t-norm, we get a conclusion where it is completely certain that the robot is at the corner. In this example, the product is a good compromise between the minimum which is cautious, and the Lukasiewicz t-norm which is adventurous.

## 4 A formal logical system (LPL) for merging possibilistic information

### 4.1 Language

In this section we will describe a pure syntactical algorithm to perform information fusion. First we will present the LPL logical language that extends the one presented for possibilistic logic, powerful enough to express from inside the logic meta-statements of the form: "combine the knowledge bases $\Sigma_1$, $\Sigma_2$ using a t-norm or a t-conorm". To this aim, we need to enrich the language with some connectives: & will represent the $min$ operator, $\oplus$ the $sup$ operator and $\otimes$ the t-norm operator. This gives the static part of the logic (i.e. combining). To have a complete logic, we need an entailment $\rightarrow$ and a negation. Moreover, all the expressive power of the preceding language must be preserved, hence, for example, we need formulae which have the following meaning: "the necessity of $A$ is greater than or equal to a number $\alpha$". To this aim, we add propositional constants $\alpha$ that will be interpreted as the possibilistic distribution with constant value $\alpha$. Lastly, we extend the possibilistic logic language to the first-order case. Summing up, we need the following language: $\mathcal{L} ::= \alpha | P | A \& B | A \oplus B | \neg A | A \rightarrow B | A \otimes B | (\forall x) A(x) | (\exists x) A(x)$. We put $\neg A =_{\text{def}} A \rightarrow 0$. We take $\mathcal{L}$ to be the set of formulae; it is convenient to define $\mathcal{L}_1$ as the set of formulae with no occurrences of $\alpha$ constants for any $\alpha \in (0, 1)$ — notice that 0 and 1 are in $\mathcal{L}_1$. We use uppercase latin letters (A, B, C,...) for formulae, while reserving L, M, N for $\mathcal{L}_1$-formulae, and uppercase greek letters ($\Gamma$, $\Delta$, ...) for multisets of formulae. As we will see, $\mathcal{L}_1$ formulae form a sub-logic which has all the properties of classical logic.

### 4.2 Semantics

As usual, to define the semantics we need a compositional function from the language to a suitable set of truth values. To find such a function we will move from the usual (global) idea of truth to a local one. Global truth means that once we have fixed a model, the truth value of a sentence is fixed, while in the local description of truth, a sentence can be true or not, depending on the available information. Hence the starting point is to fix the set of the informational states and then define when a fixed informational state makes true (forces) a formula. As the set of informational states we take the set of possibilistic distributions: $\mathcal{P}_D = \{\pi : \mathcal{M}_D \longrightarrow [0,1]\}$ where $\mathcal{M}_D$ is the set of classical first-order structures for the language $\mathcal{L}_1$ on the domain $D$ and given a sentence $L$ of the language $\mathcal{L}_1$ we will indicate with $Mod_D(L)$ the set of all structures with domain $D$ where $L$ is true. Moreover $\alpha$ is the possibilistic distribution defined as: $\alpha(w) = \alpha$, hence $\alpha$, depending on the context, can have the following meaning: the number $\alpha$, the possibilistic distribution with constant value the number $\alpha$, or the logical constant $\alpha$ (as in [Pavelka 1979]). In the following $\pi_1 \vee \pi_2$ (resp. $\pi_1 \wedge \pi_2$) corresponds to $sup(\pi_1, \pi_2)$ (resp. $inf(\pi_1, \pi_2)$), and $\leq$ represents the specificity relation. The definition of local truth is:

**Def. 4** *Let $D$ be a domain, $\times$ a t-norm. The forcing relation ($\Vdash$) between the set of possibilistic distributions and the sentences of the language $\mathcal{L}$, is defined by induction as follows:*

$$\begin{array}{rcl}
\pi \Vdash R(c_1, ..., c_n) & iff & Nec_\pi(Mod_D(R(c_1, ..., c_n))) = 1 \\
\pi \Vdash \alpha & iff & \pi \leq \alpha \\
\pi \Vdash A \otimes B & iff & (\exists \pi_1 \Vdash A)(\exists \pi_2 \Vdash B)(\pi \leq \pi_1 \times \pi_2) \\
\pi \Vdash A \oplus B & iff & (\exists \pi_1 \Vdash A)(\exists \pi_2 \Vdash B)(\pi \leq \pi_1 \vee \pi_2) \\
\pi \Vdash A \to B & iff & (\forall \pi_1 \Vdash A)(\exists \pi_2 \Vdash B)(\pi \times \pi_1 \leq \pi_2) \\
\pi \Vdash A \& B & iff & (\exists \pi_1 \Vdash A)(\exists \pi_2 \Vdash B)(\pi \leq \pi_1 \wedge \pi_2) \\
\pi \Vdash (\forall x) A(x) & iff & (\forall u \in D)(\pi \Vdash A(u)) \\
\pi \Vdash (\exists x) A(x) & iff & (\exists u \in D)(\pi \Vdash A(u))
\end{array}$$

*where $Nec_\pi : 2^{\mathcal{M}_D} \to [0,1]$ is the necessity function associated with the possibility distribution $\pi$: $Nec_\pi(X) = 1 - \bigvee_{w \notin X} \pi(w)$.*

For a given t-norm $\times$, the operator $\times : \mathcal{P}_D \times \mathcal{P}_D \longrightarrow \mathcal{P}_D$ is defined as: $(\pi_1 \times \pi_2)(w) = \pi_1(w) \times \pi_2(w)$. For any t-norm, we introduce its adjoint operation named *residuation*: $a \Rightarrow b = \bigvee\{x \in [0,1] : x \times a \leq b\}$ which is naturally extended to possibility functions: $(\pi_1 \Rightarrow \pi_2)(w) = \pi_1(w) \Rightarrow \pi_2(w)$. The negation is defined as $\neg \pi = \pi \Rightarrow 0$. The following fact is easily verified:

**Fact 4** *Any continuous t-norm $\times$ over possibility distributions distributes over infinite disjunctions, i.e: $\pi \times (\bigvee_{i \in I} \pi_i) = \bigvee_{i \in I}(\pi \times \pi_i)$ hence $Q = \langle \mathcal{P}_D, \times, 1, \leq \rangle$ is a commutative unital quantal [Rosenthal 1990]. Moreover, any t-norm $\times$ also distributes over infinite conjunctions: $\pi \times (\bigwedge_{i \in I} \pi_i) = \bigwedge_{i \in I}(\pi \times \pi_i)$*

Once we have the local concept of truth we can define the truth value of a formula $\phi$ as follows:

**Def. 5** *For every sentence $\phi$ of $\mathcal{L}$ its truth value $\|\phi\| \in Q$ is defined as: $\|\phi\| = \bigvee\{\pi \; : \; \pi \Vdash \phi\}$*

The following properties hold for the above semantics:

**Fact 5** *For any domain $D$ and continuous t-norm $\times$, the function $\|\cdot\| : \mathcal{L} \longrightarrow \mathcal{P}_D$ satisfies:*

$$\begin{array}{rcl}
\|R(c_1, ..., c_n)\| & = & \lambda w. \begin{cases} 1 & \text{if } w \models R(c_1, ..., c_n) \\ 0 & \text{otherwise} \end{cases} \\
\|\alpha\| & = & \alpha \\
\|A \& B\| & = & \|A\| \wedge \|B\| \\
\|A \oplus B\| & = & \|A\| \vee \|B\| \\
\|A \otimes B\| & = & \|A\| \times \|B\| \\
\|\forall x A(x)\| & = & \bigwedge_{u \in D} \|A(u)\| \\
\|\exists x A(x)\| & = & \bigvee_{u \in D} \|A(u)\| \\
\|A \to B\| & = & \|A\| \Rightarrow \|B\| \\
\|\neg A\| & = & \neg \|A\| \\
\|L\| & = & \lambda w. \begin{cases} 1 & \text{if } w \in Mod_D(L) \\ 0 & \text{otherwise} \end{cases} \\
\|\alpha \oplus L\| & = & \lambda w. \begin{cases} 1 & \text{if } w \in Mod_D(L) \\ \alpha & \text{otherwise} \end{cases}
\end{array}$$

The following property shows that classical logic is contained in LPL.

**Fact 6** *For any domain $D$, the set $\mathcal{B}_D = \{\pi \in \mathcal{P}_D \mid \text{for all } w \in \mathcal{M}_D, \pi(w) \in \{0, 1\}\}$ is a Boolean algebra contained in the structure $\mathcal{P}_D$.*

Note that $\mathcal{L}_1$-formulae are mapped into the Boolean algebra, so it is natural to expect that they provide an exact copy of first-order logic embedded inside $\mathcal{L}$.

### 4.3 Encoding SPL in LPL

This section shows that SPL (standard possibilistic logic) can be merged inside LPL. First, note that the last equality given in the Fact 5 suggests a natural way of representing standard possibilistic logic inside LPL: consider that the least informative (i.e., specific) possibilistic distribution $\pi$ satisfying the condition $Nec_\pi(L) \geq \alpha$ is: $\pi = \lambda w. \begin{cases} 1 & \text{if } w \models L \\ 1 - \alpha & \text{otherwise} \end{cases}$ so the token of information $Nec_\pi(L) \geq \alpha$ can be represented in LPL by $(1 - \alpha) \oplus L$. Now, it is easy to see that there is a complete translation of the aforementioned fusion operations. Assume that the set of atomic propositions is the same for the two languages: this means that the set of atomic propositions that appear in the formulae $L_i$ belonging to the base $\Sigma = \{(L_i, \alpha_i) \; : \; i = 1, \cdots, n\}$ is the same as the one that generates $\mathcal{L}$. We can give the following translation $T$ from the language of SPL to $\mathcal{L}_1$:

1. $T(P) = P$ for $P$ atomic proposition
2. $T(L \wedge M) = T(L) \& T(M)$
3. $T(L \vee M) = T(L) \oplus T(M)$
4. $T(L \to M) = T(L) \to T(M)$
5. $T(\neg L) = \neg T(L)$

Once we have a translation for the classical part we can extend it to the bases and their composition:

1. $T(\{(L_i, \alpha_i) : i = 1, \cdots, n\}) = \&_{i \leq n}(1 - \alpha_i) \oplus T(L_i)$
2. $T(\Sigma_1 \otimes \Sigma_2) = T(\Sigma_1) \otimes T(\Sigma_2)$, where $\Sigma_1 \otimes \Sigma_2$ means the combination with the t-norm of the knowledge bases $\Sigma_1, \Sigma_2$, as described in Section 3.2.
3. $T(\Sigma_1 + \Sigma_2) = T(\Sigma_1) \oplus T(\Sigma_2)$, where $+$ represents the *sup* t-conorm.

The above translation is faithful as it is shown:

**Fact 7** *For every $\Sigma$, $\pi$ is the associated distribution of $\Sigma$ iff $\pi = \|T(\Sigma)\|$.*

Hence the meta statement "fuse the two knowledge bases $\Sigma_1$ and $\Sigma_2$ using a t-norm or t-conorm operator" can be represented using a sentence of the language.

## 5 Proof system

The proof system consists of four parts: structural rules, logical rules, axioms for distributivity, and three further "numerical" axioms. The dependence of the calculus on the parameter $\times$ is quite circumscribed, and limited to numerical rules, so that the parameters do not affect the logical core of the system. The $\otimes$ connective is in general not idempotent. The calculus will therefore miss the structural rule of contraction, in the style of substructural logics [Girard 1987]; in fact a weak form of contraction is allowed, which can only be applied to the sublanguage $\mathcal{L}_1$. In presenting the calculus, we take the freedom of writing $\vdash$ instead of the parametric $\vdash_\times$, assuming we are dealing with a fixed t-norm.

**Structural rules**:

id) $\quad A \vdash A \qquad$ cut) $\quad \dfrac{\Gamma \vdash B \quad \Delta, B \vdash C}{\Delta, \Gamma \vdash C}$

exL) $\quad \dfrac{\Gamma, B, A, \Delta \vdash C}{\Gamma, A, B, \Delta \vdash C} \qquad$ weL) $\quad \dfrac{\Gamma \vdash B}{\Gamma, A \vdash B}$

$\mathcal{L}_1$con) $\quad \dfrac{\Gamma, A, L \vdash B \quad \Delta, A \vdash L}{\Gamma, \Delta, A \vdash B} \quad L \in \mathcal{L}_1$

**Logical rules**:

&) $\quad \dfrac{\Gamma, A \vdash C}{\Gamma, A \& B \vdash C} \quad \dfrac{\Gamma, B \vdash C}{\Gamma, A \& B \vdash C} \qquad \dfrac{\Gamma \vdash A \quad \Gamma \vdash B}{\Gamma \vdash A \& B}$

$\otimes$) $\quad \dfrac{\Gamma, A, B \vdash C}{\Gamma, A \otimes B \vdash C} \qquad \dfrac{\Gamma \vdash A \quad \Delta \vdash B}{\Gamma, \Delta \vdash A \otimes B}$

$\oplus$) $\quad \dfrac{\Gamma, A \vdash C \quad \Gamma, B \vdash C}{\Gamma, B \oplus A \vdash C} \qquad \dfrac{\Gamma \vdash A}{\Gamma \vdash A \oplus B} \quad \dfrac{\Gamma \vdash B}{\Gamma \vdash A \oplus B}$

$\rightarrow$) $\quad \dfrac{\Gamma \vdash A \quad \Delta, B \vdash C}{\Gamma, \Delta, A \rightarrow B \vdash C} \qquad \dfrac{\Gamma, A \vdash B}{\Gamma \vdash A \rightarrow B}$

$\forall$) $\quad \dfrac{\Gamma, A(t) \vdash B}{\Gamma, \forall x A(x) \vdash B} \qquad \dfrac{\Gamma \vdash A(x)}{\Gamma \vdash \forall x A(x)}*$

$\exists$) $\quad \dfrac{\Gamma, A(x) \vdash B}{\Gamma, \exists x A(x) \vdash B}* \qquad \dfrac{\Gamma \vdash A(t)}{\Gamma \vdash \exists x A(x)}$

1) $\quad \dfrac{\Gamma \vdash A}{\Gamma, 1 \vdash A} \qquad \Gamma \vdash 1$

0) $\quad \Gamma, 0 \vdash A$

$\neg\neg$) $\quad \neg\neg L \vdash L \quad L \in \mathcal{L}_1$

\* $x$ not free in $\Gamma, B$

**Distributivity**:

$\otimes - \&$ distr) $\quad (A \otimes C) \& (B \otimes C) \vdash (A \& B) \otimes C$
$\oplus - \&$ distr) $\quad (A \oplus C) \& (B \oplus C) \vdash (A \& B) \oplus C$
$\otimes - \forall$ distr) $\quad \forall x A(x) \otimes C \dashv\vdash \forall x (A(x) \otimes C) \quad$ if $x$ is not free in $C$
CD) $\quad \forall x (A \oplus B(x)) \vdash A \oplus \forall x B(x) \quad$ if $x$ is not free in $A$

**Numerical rules**:

S') $\quad \beta \vdash \alpha \qquad$ for any $\beta \leq \alpha$
$\otimes$ def) $\quad \alpha \otimes \beta \dashv\vdash \gamma \qquad$ where $\gamma = \alpha \times \beta$
$\neg$def) $\quad \neg\alpha \dashv\vdash \gamma \qquad$ where $\gamma = \alpha \Rightarrow 0$

Validity and completeness hold, as the following theorem shows:

**Theorem 2** *For any $\Gamma$ and $B$: $\Gamma \vdash B$ if and only if for every $\pi$ if $\pi \Vdash \otimes \Gamma$ then $\pi \Vdash B$.*

where if $\Gamma$ is the multiset $\{A_1, \cdots, A_n\}$ then $\otimes \Gamma =_{df} A_1 \otimes A_2 \otimes \cdots \otimes A_n$. All the proofs are in [Boldrin and Sossai 1997]. Let us recall the main problem: we have some amount of information represented by some graded formulae, we apply the fusion procedure and obtain a theory $\Sigma$. Given a formula $A$ what is the degree of necessity of $A$ computed using the available knowledge? If $\Sigma \vdash \alpha \oplus A$ then, due to the validity property of the proof system, for every $\pi$ s.t. $\pi \Vdash \Sigma$ we have that $\pi \Vdash \alpha \oplus A$ and this means that: $Nec_\pi(A) \geq 1 - \alpha$. On the other hand, completeness gives the following property: if for every $\pi \Vdash \Sigma$ we have that $Nec_\pi(A) \geq 1 - \alpha$ then there is a proof of the sequent: $\Sigma \vdash \alpha \oplus A$. Hence the proof system gives a correct (validity) and powerful enough (completeness) procedure to solve the problem. Note that every proof of $\Gamma \vdash \alpha \oplus A$ terminates with two types of leaves:

1. purely logical, i.e. of the form $L \vdash L$
2. purely numerical, i.e. $\beta \vdash \alpha$, or $\alpha \otimes \beta \vdash \gamma$, ...

The purely numerical leaves give the constraints which $\beta$, $\alpha$, $\gamma$ must satisfy: for example, $\beta \vdash \alpha$ is provable if and only if $\beta \leq \alpha$. Hence we can use the sequent calculus to determine the necessity value of a formula $A$ as follows: start searching a proof of $x \oplus A$ with $x$ undetermined, such proof gives the constraints which determine the value of $x$.

### 5.1 Example

Let us consider again the robotics example. First the available information, given in the example of Section 3.3., can be codified in our language using the following set $\Gamma : ((D \& E \& F) \oplus G) \rightarrow H, \eta \oplus ((A \& B) \rightarrow D), \theta_1 \oplus A, \theta_2 \oplus B, \theta_4 \oplus E, \theta_5 \oplus F, \theta_6 \oplus G$. Now let us see how a proof (i.e. a single sequent) of $\Gamma \vdash x \oplus H$ allows us to compute the value of $x$ as a function of the degrees of the formulae appearing in $\Gamma$.

Due to the lack of space we will split the sequent into some smaller parts, nevertheless the reader can easily reconstruct the complete proof. We will start with three sub-proofs that appear inside the proof of $\Gamma \vdash x \oplus H$.

&-Comp) Note that the following are provable:

1.
$$\frac{\frac{\frac{\alpha_1 \vdash x}{\alpha_1 \vdash x}}{\frac{\alpha_1 \vdash x \oplus (L_1 \& L_2)}{\alpha_1, \neg L_1 \& \neg L_2 \vdash x \oplus (L_1 \& L_2)}} \quad \frac{\frac{L_1 \vdash L_1}{L_1 \vdash L_1, x \oplus (L_1 \& L_2)}}{\frac{L_1, \neg L_1 \vdash x \oplus (L_1 \& L_2)}{L_1, \neg L_1 \& \neg L_2 \vdash x \oplus (L_1 \& L_2)}}}{\frac{(\alpha_1 \oplus L_1), \neg L_1 \& \neg L_2 \vdash x \oplus (L_1 \& L_2)}{(\alpha_1 \oplus L_1) \& (\alpha_2 \oplus L_2), \neg L_1 \& \neg L_2 \vdash x \oplus (L_1 \& L_2)}}$$

Note that this is a proof iff $\alpha_1 \vdash x$ is provable i.e. iff $\alpha_1 \leq x$

2.
$$\frac{\frac{\frac{\alpha_2 \vdash x}{\alpha_2 \vdash x \oplus (L_1 \& L_2)}}{\alpha_2, \neg L_2 \& \neg L_2 \vdash x \oplus (L_1 \& L_2)} \quad \frac{\frac{L_2 \vdash L_2}{L_2, \neg L_2 \vdash 0} \quad 0 \vdash x \oplus (L_1 \& L_2)}{\frac{L_2, \neg L_2 \vdash x \oplus (L_1 \& L_2)}{L_2, \neg L_1 \& \neg L_2 \vdash x \oplus (L_1 \& L_2)}}}{\frac{(\alpha_2 \oplus L_2), L_1 \& \neg L_2 \vdash x \oplus (L_1 \& L_2)}{(\alpha_1 \oplus L_1) \& (\alpha_2 \oplus L_2), L_1 \& \neg L_2 \vdash x \oplus (L_1 \& L_2)}}$$

Note that this is a proof iff $\alpha_2 \vdash x$ is provable i.e. iff $\alpha_2 \leq x$

3. The case: $(\alpha_1 \oplus L_1) \& (\alpha_2 \oplus L_2), \neg L_1 \& L_2 \vdash x \oplus (L_1 \& L_2)$ is similar to the above one.

4.
$$\frac{\frac{L_1 \& L_2 \vdash L_1 \& L_2}{L_1 \& L_2 \vdash x \oplus (L_1 \& L_2)}}{(\alpha_1 \oplus L_1) \& (\alpha_2 \oplus L_2), L_1 \& L_2 \vdash x \oplus (L_1 \& L_2)}$$

No constraints on $x$

All the above sequents are provable iff $x \geq \alpha_1$ and $x \geq \alpha_2$, i.e. $x \geq (\alpha_1 \vee \alpha_2)$. If we apply the $\oplus$-rule on the left to the four proofs we obtain the sequent: $(\alpha_1 \oplus L_1) \& (\alpha_2 \oplus L_2), (\neg L_1 \& \neg L_2) \oplus (L_1 \& \neg L_2) \oplus (\neg L_1 \& L_2) \oplus (L_1 \& L_2) \vdash (\alpha_1 \oplus \alpha_2) \oplus (L_1 \& L_2)$. Note that due to the fact that $(\neg L_1 \& \neg L_2) \oplus (L_1 \& \neg L_2) \oplus (\neg L_1 \& L_2) \oplus (L_1 \& L_2)$ is a classical tautology we have that $\vdash (\neg L_1 \& \neg L_2) \oplus (L_1 \& \neg L_2) \oplus (\neg L_1 \& L_2) \oplus (L_1 \& L_2)$ is a provable sequent and hence, due to the fact that weakening holds, $(\alpha_1 \oplus L_1) \& (\alpha_2 \oplus L_2) \vdash (\neg L_1 \& \neg L_2) \oplus (L_1 \& \neg L_2) \oplus (\neg L_1 \& L_2) \oplus (L_1 \& L_2)$ is a provable sequent. Now using $\mathcal{L}_1$ contraction we get a proof of: $(\alpha_1 \oplus L_1) \& (\alpha_2 \oplus L_2) \vdash x \oplus (L_1 \& L_2)$ iff $x \geq \alpha_1 \vee \alpha_2$
The following gives us a Generalized Modus Ponens

**GM)**

$$\frac{\frac{\beta \vdash x}{\beta \vdash x \oplus B}}{\beta, \alpha \oplus (A \to B) \vdash x \oplus B} \quad \frac{\frac{\alpha \vdash x}{\alpha \vdash x \oplus B} \quad \frac{A \vdash A \quad \frac{B \vdash B}{B \vdash x \oplus B}}{A, A \to B \vdash x \oplus B}}{\frac{A, \alpha \oplus (A \to B) \vdash x \oplus B}{\alpha \oplus (A \to B), \beta \oplus A \vdash x \oplus B}}$$

Note that: $\alpha \oplus (A \to B), \beta \oplus A \vdash x \oplus B$ iff $x \geq \alpha \vee \beta$

⊗-Comp)

$$\frac{\frac{\xi_2 \vdash \xi_2 \quad \xi_1 \vdash \xi_1}{\xi_2, \xi_1 \vdash \xi_2 \otimes \xi_1} \quad \xi_2 \otimes \xi_1 \vdash x}{\frac{\xi_2, \xi_1 \vdash x}{\frac{\xi_2, \xi_1 \vdash x \oplus H}{(K \oplus G) \to H, \xi_2, \xi_1 \vdash x \oplus H}}} \quad \frac{\frac{G \vdash G}{G \vdash K \oplus G} \quad H \vdash H}{\frac{(K \oplus G) \to H, G \vdash H}{\frac{(K \oplus G) \to H, G \vdash x \oplus H}{(K \oplus G) \to H, \xi_2, G \vdash x \oplus H}}}}{(K \oplus G) \to H, K, \xi_1 \oplus G \vdash x \oplus H}$$

The above is a proof iff $x \geq \xi_1 \times \xi_2$

$$\frac{\frac{K \vdash K}{K \vdash K \oplus G} \quad H \vdash H}{\frac{(K \oplus G) \to H, K \vdash H}{\frac{(K \oplus G) \to H, K \vdash x \oplus H}{(K \oplus G) \to H, K, \xi_1 \vdash x \oplus H}}} \quad \frac{\frac{K \vdash K}{K, G \vdash K}}{\frac{K, G \vdash K \oplus G \quad H \vdash H}{\frac{(K \oplus G) \to H, K, G \vdash H}{(K \oplus G) \to H, K, G \vdash x \oplus H}}}$$
$$(K \oplus G) \to H, K, \xi_1 \oplus G \vdash x \oplus H$$

No constraints on $x$. Combining the two above proofs with an $\oplus$-rule and then applying a $\otimes$ rule we obtain the following: $(K \oplus G) \to H, (\xi_2 \oplus K) \otimes (\xi_1 \oplus G) \vdash x \oplus H$ iff $x \geq \xi_1 \times \xi_2$ Due to the lack of space we must assume the following abbreviation:

1. $\Gamma_1 : (\theta_4 \oplus E) \& (\theta_5 \oplus F), ((D \& E \& F) \oplus G) \to H, \eta \oplus ((A \& B) \to D), (\theta_1 \oplus A) \& (\theta_2 \oplus B), \theta_6 \oplus G$.

2. $\Gamma_2 : ((D \& E \& F) \oplus G) \to H, (\xi_2 \oplus (D \& E \& F)) \otimes (\theta_6 \oplus G)$.

3. $\Gamma_3 : (\theta_4 \oplus E) \& (\theta_5 \oplus F), \eta \oplus ((A \& B) \to D), (\theta_1 \oplus A) \& (\theta_2 \oplus B), \theta_6 \oplus G$.

4. $\Gamma_4 : (\theta_4 \oplus E) \& (\theta_5 \oplus F), \eta \oplus ((A \& B) \to D), (\theta_1 \oplus A) \& (\theta_2 \oplus B)$.

Let us prove $\Gamma_1 \vdash x \oplus H$:

$$\frac{\frac{\Gamma_4 \vdash \xi_2 \oplus (E \& F \& D) \quad \theta_6 \oplus G \vdash \theta_6 \oplus G}{\Gamma_3 \vdash (\theta_6 \oplus G) \otimes (\xi_2 \oplus (E \& F \& D))} \otimes - Comp \quad \Gamma_2 \vdash x \oplus H}{\Gamma_1 \vdash x \oplus H}$$

Let us prove $\Gamma_4 \vdash \xi_2 \oplus (E \& F \& D)$

$$\frac{\Gamma_4 \vdash (\alpha \oplus (E \& F)) \& (\beta \oplus D) \quad \frac{\& - Comp_1}{(\alpha \oplus (E \& F)) \& (\beta \oplus D) \vdash \xi_2 \oplus (E \& F \& D)}}{\Gamma_4 \vdash \xi_2 \oplus (E \& F \& D)}$$

The proof of $\Gamma_4 \vdash (\alpha \oplus (E \& F)) \& (\beta \oplus D)$ is:

$$\frac{\frac{\& - Comp_3}{(\theta_4 \oplus E) \& (\theta_5 \oplus F) \vdash \alpha \oplus (E \& F)}}{\Gamma_4 \vdash \alpha \oplus (E \& F)}$$

$$\frac{\frac{\& - Comp_2}{(\theta_1 \oplus A) \& (\theta_2 \oplus B) \vdash \gamma \oplus (A \& B)} \quad \frac{GM}{\eta \oplus (A \& B \to D), \gamma \oplus (A \& B) \vdash \beta \oplus D}}{\frac{(\theta_1 \oplus A) \& (\theta_2 \oplus B), \eta \oplus (A \& B \to D) \vdash \beta \oplus D}{\Gamma_4 \vdash \beta \oplus D}}$$

now applying an & rule to the two above proven formulas we obtain: $\Gamma_4 \vdash (\alpha \oplus (E \& F)) \& (\beta \oplus D)$ Note that the above proof gives us the following constraints:

1. from $\otimes - Comp$ we have that: $x \geq \theta_6 \times \xi_2$.
2. from $\& - Comp_1$ we have that: $\xi_2 \geq \alpha \vee \beta$
3. from $GM$ we have that: $\beta \geq \gamma \vee \eta$.
4. from $\& - Comp_2$ we have that: $\gamma \geq \theta_1 \vee \theta_2$
5. from $\& - Comp_3$ we have that: $\alpha \geq \theta_4 \vee \theta_5$

Solving all the above inequalities we have: $x \geq \theta_6 \times (\theta_1 \vee \theta_2 \vee \theta_4 \vee \theta_5 \vee \eta)$ If we want the highest value for the necessity of $H$ compatible with the above constraints we get the following function (which leads to minimize the value of x): $x = \theta_6 \times (\theta_1 \vee \theta_2 \vee \theta_4 \vee \theta_5 \vee \eta)$ If we substitute the numbers to the corresponding variables, the above function gives us the same number as in the third proposed solution to the example of Section 3.3. In fact, $N(H) = 1 - x = 1 - 0.4 \times (0.6 \vee 0.5 \vee 0.3 \vee 0.6 \vee 0.2) = 1 - 0.4 \times 0.6 = 1 - 0.24 = 0.76$. This procedure computes $N(H)$ as a function of the available information. It is also interesting to note that $\Gamma_1 : (\theta_4 \oplus E)\&(\theta_5 \oplus F), ((D\&E\&F) \oplus G) \to H, \eta \oplus ((A\&B) \to D), (\theta_1 \oplus A)\&(\theta_2 \oplus B), \theta_6 \oplus G$ gives us an explicit analysis of the implicit dependency/independency assumptions made in the computation of $x$. In fact a comma inside $\Gamma_1$ is equivalent to a $\otimes$-combination of the formulae, hence the formulae separated by a comma are assumed as independent, while those combined by the & are assumed as not independent: for example $(\theta_4 \oplus E)\&(\theta_5 \oplus F)$ and $((D\&E\&F) \oplus G) \to H$ are assumed as independent, while $(\theta_4 \oplus E)$ and $(\theta_5 \oplus F)$ are assumed as dependent. Last but not least, let us stress the fact that here logic is used as a mathematical method to study the function that computes the degree of the conclusion using the degrees of the premises. Such function is directly implemented on the robot.

## 6   Conclusions

This paper has proposed several syntactical combination modes which can be applied to possibilistic knowledge bases. The work presented in this paper is closely related to the one proposed in (Boldrin and Saffiotti, 1995). Several differences can be mentioned: first their work is based on possible world semantics while in this paper the algebraic one is used. Secondly, in their world it is possible to identify the different sources (using one modality per source) which is not possible in our work. Thirdly, negation used in the two logics is not the same. Lastly, in this paper it is possible to have a many-level merging of information using diferent t-norms which is not possible in their approach.

Two methods have been investigated: the first one is achieved at the standard possibilistic logic while the second method is based on an extension of the language of possibilistic logic. The first method requires a preliminary partitioning of the knowledge base into sub-bases, then the combination of the sub-bases to generate new knowledge bases, and eventually the application of the resolution algorithm to calculate the necessity degree of a formula. In the second method this process is locally controlled by the sequent calculus, which chooses in the base the formulae required for the proof, and it combines them as appropriate to the development of the demonstration. The main limit of the first method is that it does not treat predicate logic, while the main limit of the second method is that it does not allow the combination of bases with t-conorms different from the maximum. Lastly, the combination techniques proposed in this paper have been successfully applied to the problem of position estimation (localization) of a mobile robot in a partially and approximately known indoor environment (for more details, see [Bison et al., 1997a,b,1998]).

### Acknowledgments
This work was partially supported by a CNR-CNRS joint project (CNR Code 132.3.1) and the Fusion Project of the European Community. The second author is indebted to A. Saffiotti of Iridia and to G. Chemello of Ladseb.

## References

[Abidi and Gonzalez 1992] Abidi M.A. and Gonzalez R.C. Data Fusion in Robotics and Machine Intelligence. Academic Press, New York.

[Benferhat et al. 1995] Benferhat S., Dubois D. and Prade H. How to infer from inconsistent beliefs without revising?. Proc. of the 14th Inter. Joint Conf. on Artif. Intell. (IJCAI'95), 1449-1455.

[Benferhat et al. 1997] Benferhat S., Dubois D. and Prade H. From semantic to syntactic approaches to information combination in possibilistic logic". In Aggregation and Fusion of Imperfect Information, Studies in Fuzziness and Soft Computing (B. Bouchon-Meunier, ed.), Physica Verlag, 141-151.

[Bison et al. 1997a] , P. Bison, G. Chemello, C. Sossai and G. Trainito, Logic-Based Sensor Fusion for Localization, in *Proc. of the IEEE International Symposium on Computational Intelligence in Robotics and Automation (CIRA'97)*, IEEE CS, pp. 254-259.

[Bison et al. 1997b] , P. Bison, G. Chemello, C. Sossai and G. Trainito, A Possibilistic Approach to Sensor Fusion in Mobile Robotics, in *Proceedings of the 2nd Euromicro Workshop on Advanced Mobile Robots (Eurobot'97)*, IEEE Computer Society, pp. 73-79.

[Bison et al. 1998] , P. Bison, G. Chemello, C. Sossai and G. Trainito, Cooperative Localization using Possibilistic Sensor Fusion, in *Proc. of the 3rd IFAC Symposium on Intell. Autonomous Vehicles (IAV'98)*.

[Boldrin and Saffiotti 1995] L. Boldrin and C. Saffiotti. A modal logic for merging partial belief of multiple reasoners. Technical Report TR/IRIDIA/95-19.

[Boldrin and Sossai 1995] L. Boldrin and C. Sossai. An algebraic semantics for possibilistic logic. In Proc. of UAI-95, pages 24-37.

[Boldrin and Sossai 1997] L. Boldrin and C. Sossai. Local Possibilistic Logic, *Journal of Applied Non-Classical Logic*, 7(3):309-333, 1997.

[Cholvy1992] A logical approach to multi-sources reasoning. In: Applied Logic Conference: Logic at Work, Amsterdam.

[Dubois et al. 1994] D. Dubois, J. Lang and H. Prade. Possibilistic logic. In D. M. Gabbay et al., eds, *Handbook of Logic in Artificial Intelligence and Logic Programming (Volume 3)*, 439–513.

[Dubois and Prade 1992] Dubois D., Prade H. Combination of fuzzy information in the framework of possibility theory. In: Data Fusion in Robotics and Machine Intelligence (M.A. Abidi, R.C. Gonzalez, eds.) Academic Press, New York, 481-505.

[Flamm and Luisi 1992] Flamm and Luisi. Reliability Data and Analysis. Kluwer Academic Publ.

[Girard 1987] J.-Y. Girard. Linear logic. *Theoretical computer science*, 50:1–101, 1987.

[Pavelka 1979] J. Pavelka. On fuzzy logic i, ii, iii. *Zeitschr. f. math. Logik und Grundlagen d. Math*, 25:45–52; 119–131; 447–464, 1979.

[Rosenthal 1990] K. I. Rosenthal. *Quantales and their applications*. Longman, 1990.